\documentclass{article}

\usepackage[final]{cml4impact_2022}

\usepackage[utf8]{inputenc} % allow utf-8 input
\usepackage[T1]{fontenc}    % use 8-bit T1 fonts
\usepackage{hyperref}       % hyperlinks
\usepackage{url}            % simple URL typesetting
\usepackage{booktabs}       % professional-quality tables
\usepackage{amsfonts}       % blackboard math symbols
\usepackage{nicefrac}       % compact symbols for 1/2, etc.
\usepackage{microtype}      % microtypography
\usepackage{xcolor}         % colors
\usepackage{graphicx}
\usepackage{float}
\usepackage{subcaption}
\usepackage{wrapfig}
\usepackage{tabto}
\usepackage[ruled, lined, linesnumbered, commentsnumbered, noend]{algorithm2e}
\title{Improving the Efficiency of the PC Algorithm by Using Model-Based Conditional Independence Tests}

\author{%
  Erica Cai, Andrew McGregor, David Jensen \\
  Manning College of Information \& Computer Sciences\\
  University of Massachusetts Amherst\\
  \texttt{\{ecai,mcgregor,jensen\}@cs.umass.edu} \\
}

\begin{document}

\maketitle

\begin{abstract}
Learning causal structure is useful in many areas of artificial intelligence, including planning, robotics, and explanation. Constraint-based structure learning algorithms such as PC use conditional independence (CI) tests to infer causal structure. Traditionally, constraint-based algorithms perform CI tests with a preference for smaller-sized conditioning sets, partially because the statistical power of conventional CI tests declines rapidly as the size of the conditioning set increases. However, many modern conditional independence tests are \textit{model-based}, and these tests use well-regularized models that maintain statistical power even with very large conditioning sets. This suggests an intriguing new strategy for constraint-based algorithms which may result in a reduction of the total number of CI tests performed: Test variable pairs with \textit{large} conditioning sets \textit{first}, as a pre-processing step that finds some conditional independencies quickly, before moving on to the more conventional strategy that favors small conditioning sets. We propose such a pre-processing step for the PC algorithm which relies on performing CI tests on a few randomly selected large conditioning sets. We perform an empirical analysis on directed acyclic graphs (DAGs) that correspond to real-world systems and both empirical and theoretical analyses for Erd\H{o}s-Renyi DAGs. Our results show that Pre-Processing Plus PC (P3PC) performs far fewer CI tests than the original PC algorithm, between 0.5\% to 36\%, and often less than 10\%, of the CI tests that the PC algorithm alone performs. The efficiency gains are particularly significant for the DAGs corresponding to real-world systems.
% We perform experiments on directed acyclic graphs (DAGs) that correspond to real-world systems and on Erd\H{o}s-Renyi DAGs, and show that the PC algorithm with pre-processing performs far fewer CI tests than the original PC algorithm, between 0.5\% and 20\%, of the CI tests that the PC algorithm alone performs. The efficiency gains are particularly large on DAGs that correspond to real-world systems, 

% 

%CML4im
\end{abstract}

\section{Introduction}
Causal models are useful for representing and reasoning about causal relationships in many areas of artificial intelligence and machine learning. One way to represent causal models is as a directed graphical model, which consists of a causal structure in the form of directed edges between variables and parameters in the form of a conditional probability distribution corresponding to each variable. The edges in the directed structure convey the existence and direction of causal relationships, and the parameters specify the magnitude of their causal effects. Some approaches to constructing a causal graphical model first learn a graph structure and then use that graph structure for estimating parameters.

Constraint-based algorithms for structure learning infer a structure that corresponds to the conditional independence (CI) relationships within a population. These conditional independence relationships are inferred by performing CI tests on a data sample.  In this work, we focus on constraint-based algorithms, improving their efficiency for causal structure learning.

The phase of constraint-based algorithms that applies CI tests typically dominates the runtime of such algorithms, and the number of such tests in the worst case is equal to the cardinality of the power set of the possible conditioning variables. Specifically, constraint-based algorithms run a CI test for each pair of variables and possible conditioning set, stopping only when a test infers that the pair is conditionally independent or after all feasible CI tests have been performed. As the number of variables increases, the number of CI tests performed by these algorithms increases exponentially. 

Most constraint-based algorithms assume that CI tests perform more accurately on small conditioning sets. However, recent model-based CI tests have the potential to work effectively on very large conditioning sets. The models typically used in those tests are well-regularized, and so they can produce accurate results even with very large numbers of conditioning variables. Several experimental evaluations have found that model-based tests perform more accurately than older CI tests, particularly on larger conditioning sets. Substantial recent work on CI tests suggests that their accuracy and efficiency is likely to improve. 

The focus of this paper is on how such current and future tests could be used to improve the efficiency of constraint-based algorithms for structure learning. We assume that accurate and general model-based CI tests exist or will exist soon. We explore how algorithms could make use of model-based CI tests, presenting a pre-processing algorithm for the PC algorithm which relies on randomly selecting a few large conditioning sets and running model-based CI tests on those sets for every pair of variables. The aim of the pre-processing algorithm is to speed up the process of identifying pairs of variables that are conditionally independent. We provide theoretical and empirical evidence that the pre-processing algorithm is extremely successful at this task.

The core idea that makes the pre-processing algorithm useful is to perform CI tests on \textit{large} conditioning sets. Large conditioning sets are often avoided when designing algorithms for structure learning due to a concern that such sets could easily include one or more colliders. Conditioning on a collider can induce dependence between a pair of variables that would otherwise be independent. However, we show that the probability of conditioning on a collider in an otherwise unblocked path is very small when the size of the conditioning set is large. This enables conditioning on large variable sets to very quickly identify large numbers of conditional independence relations.

We show that induced dependence due to conditioning on a collider is very rare when conditioning sets are large. Further, we observe that failing to correctly infer \textit{every} conditional independence in the pre-processing step is not a major problem, because unrecognized CIs will be caught in the next phase of the algorithm, and that finding many CIs early greatly reduces the total number of tests. We conduct a theoretical analysis and an experimental evaluation to explore the efficiency gain between the PC algorithm with pre-processing and the PC algorithm without such pre-processing. We conduct the experiments on both random Erd\H{o}s-Renyi (ER) directed acyclic graphs (DAGs) and DAGs that correspond to real-world systems. We find that Pre-Processing Plus PC (P3PC) performs between 0.5\% to 36\%, and often less than 10\%, of the CI tests that the PC algorithm alone performs on DAGs that correspond to real-world systems. 

\section{Related Work}

A wide variety of work has focused on how to improve the accuracy and efficiency of structure learning algorithms, including work on constraint-based algorithms \cite{bromberg2014efficient}, score-based algorithms, continuous-optimization algorithms \cite{vowels2022dya}, and combinations of these approaches \cite{tsamardinos2006maxmin}. Advances in accuracy and efficiency for continuous-optimization algorithms have been particularly prominent recently \cite{zheng2018dags}, turning some structure learning into combinatorial optimization and convex optimization problems \cite{lee2006efficientstructure}. Other advances in efficiency have included analyzing properties of scoring criteria such as BIC and AIC \cite{zhu2021efficient}.

In contrast, in this paper we focus on improving the efficiency of constraint-based algorithms, particularly the PC algorithm \cite{spirtes1993causation}. The PC algorithm performs CI tests in order of increasing size of conditioning sets (0, 1, 2, \ldots). The algorithm removes an edge from a complete undirected graph once it discovers that a conditioning set makes the corresponding pair of variables conditionally independent. Several avenues exist for improving the efficiency of the PC algorithm. First, some work has focused on reducing the number of CI tests that need to be performed \cite{tsamardinos2003algorithms, zhang2021fast, sondhi2019reduced, giudice2021dual, QI2021mutual}. Second, some work has focused on improving the PC algorithm in specific applications where the general algorithm performance is weaker, such as for partially faithful distributions in high-dimensional linear models, when the data contains outliers, and when the data is ordinal \cite{musella2013pc, kalisch2008robustification, wien2021approach, Buhlmann2010VariableSI}. Third, some work has focused on improving efficiency through the use of GPUs and parallelization \cite{hagedorn2021gpu, scutari2017bayesian, le2015fast, zhang2021fast, Marbach2012Wisdom, madsen2015parallel, madsen2016parallel, schmidt2018order}. 

However, all of the above techniques for constraint-based algorithms rely on the fact that the PC algorithm must prioritize performing CI tests with small conditioning sets. The literature suggests that the older CI tests do not perform well on larger conditioning sets and that mistakes early on in the PC algorithm severely affect the structure learned by the PC algorithm since early errors propagate forward \cite{spirtes1993causation}. 

Although older CI tests are typically inaccurate given large conditioning sets, new conditional independence tests continue to be developed. Some new tests are model-based \cite{kubkowski2021how}, and perform far better on large conditioning sets. One example is the Classifier Conditional Independence Test (CCIT), which constructs classification models using the conditioning set and other variables to test conditional independence \cite{sen2017model}. Modern CI tests such as CCIT and related methods are far less sensitive to the size of the conditioning set than more conventional tests. Further, many CI tests are fast and are becoming faster. Some very fast CI tests such as the Randomized Conditional Independence Test (RCIT) \cite{strobl2017approx} are improvements to CI tests that were too slow to be applied such as the Kernel-Based Conditional Independence Test (KCIT) \cite{zhang2011kernel}. Other CI tests such as CCIT are not almost instantaneous, but also not intractable. Given these new improvements and work, we could assume that even if a CI test does not meet a standard of quickness and accuracy, such a test will be developed in the near future.

\section{Pre-processing Algorithm}

We propose a pre-processing algorithm for the PC algorithm that leverages the use of CI tests which perform accurately and efficiently on large conditioning sets:

\begin{algorithm}[H]
\DontPrintSemicolon
\SetAlgoLined
\SetKwInOut{Input}{Input}
\SetKwInOut{Output}{Output}

\Input{$X=[\textbf{x}_1,\textbf{x}_2,\ldots,\textbf{x}_n]$: a dataset where $\textbf{x}_i$ is a vector containing data samples for the $i$th variable and $n$ is the number of variables in the dataset; $V$: the set of variables}
\Output{$p=[v_1,v_2,\ldots,v_n]$: an $n\times n$ matrix where index $[i,j]$ is $0$ if the pair of variables could be made conditionally independent and $1$ otherwise}
	Initialize $p$ as an $n\times n$ matrix with all $1$ values. \; Initialize $L$ as a list of sets having length $c_1$. \;
	\For{$a,b\in V$} {
			\If{$a \perp b$}{Set $p[a,b]=0$ and $p[b,a]=0$; go to the next iteration of the for loop\;}
				\For{$i \in \{1, \ldots, c_1\}$} {Assign $L[i]$ with a random set of $n-c_2$ variables not including $a$ or $b$\;
					%Append a random set of $n-c_2$ variables not including $a$ and $b$ to list $L$\;
				}
			\For{$S\in L$} {
				\If{$a \perp b \mid S$}{Set $p[a,b]=0$ and $p[b,a]=0$; go to the next iteration of the for loop\;}
			
	}}
	Return $p$\;
	
 \caption{Pre-processing algorithm for the PC algorithm}
\end{algorithm}

The pre-processing algorithm can be used to inform the PC algorithm about conditionally independent variable pairs before PC runs. Such knowledge decreases the total number of CI tests that PC performs (see details in Sections 4 and 5). For each pair of variables, the algorithm randomly selects $c_1$ conditioning sets of size $n-c_2$ where $c_1$ and $c_2$ are both small positive integers. For the derivations in Section 4, we set $c_1=3$ and $c_2=4$. Next, the algorithm performs a CI test using the empty set and on each of the $c_1$ conditioning sets to determine whether the variable pair is conditionally independent. 

The pre-processing algorithm we propose relies on tests that are accurate on large conditioning sets (e.g., a model-based CI test). However, after pre-processing, the subsequent application of the PC algorithm still uses traditional CI tests. Although fast model-based CI tests exist, given the current relative speed of model-based and traditional CI tests, the most efficient algorithm is likely to use model-based CI tests only for pre-processing.

\section{Theoretical Analysis for Erd\H{o}s-Renyi DAGs}

\textbf{Definitions.} We define a \textit{trail} of length $\ell$ between nodes $u_0$ and $u_{\ell}$ as a sequence of $\ell$ distinct edges $e_1, \ldots, e_\ell$ where $u_0$ is an endpoint of $e_1$, $u_\ell$ is an endpoint of $e_\ell$, and each $e_i$ has one endpoint in common with $e_{i-1}$ and the other endpoint in common with $e_{i+1}$. Note that the head (or ``child'' node) of $e_{i-1}$ does not need to equal the tail (or ``parent'' node) of $e_i$.
% shares  of the form $\{u_0,u_1\}, \{u_1, u_2\}, \ldots, \{u_{\ell-1},u_\ell\}$.  
The node in common with $e_i$ and $e_{i+1}$ is a \textit{collider} in the trail if it is the head of both the edges $e_i$ and $e_{i+1}$.
% A \textit{collider} of a trail is a node on the trail such that one parent is to the direct left of it on the trail and the other parent is to the direct right of it on the trail. 
A trail is \textit{blocked} by a conditioning set if at least one intermediate node on the trail that is not a collider \textit{is} in the conditioning set or if at least one intermediate node on the trail that is a collider is \textit{not} in the conditioning set. A \textit{pair of variables is inferred to be conditionally independent} if all trails between the pair of variables are blocked. The goal of the pre-processing algorithm is to detect whether a pair of variables can be inferred to be conditionally independent. 

\textbf{Erd\H{o}s-Renyi DAGs.} An Erd\H{o}s-Renyi (ER) DAG on $n$ nodes is randomly generated as follows: the $n$ nodes are labeled in increasing order and an edge exists with probability $p$ and can only have a direction from nodes with a smaller label to nodes with a larger label. For ER DAGS, we provide indications that one randomly selected conditioning set of size $n-4$ would be able to determine whether a pair of variables is CI. We provide the theoretical analysis with respect to conditioning sets specifically of size $n-4$ despite being able to generalize the result to conditioning sets of size $n-c_2$ for $2\leq c_2\leq n$ because $n-4$ is guaranteed to be a reasonably large conditioning set size for most DAGs. Thus, we could more easily examine the effect of a \textit{large} conditioning set size on the process of determining whether a pair of variables could possibly be made CI. We index the nodes $v_1,v_2,\ldots,v_n$ where $v_i$ is the node with the $i$th smallest label.

If a pair of variables could be detected as CI with respect to a conditioning set of size $n-4$, then all possible trails between the pair of variables are blocked with respect to the conditioning set. We indicate that the probability that all possible trails between a pair of variables are blocked with respect to the conditioning set is low by considering trails of length $7$ and more and trails of length $6$ and fewer separately. The specific raw values of $7$ and $6$ rely on the fixed large conditioning set size $n-4$. 

\textbf{Statement 1 (Trails of length at least $7$).} 
A conditioning set of size $n-4$ will block every trail of length $7$ or greater. Note that this will be true for any DAG, not just Erd\H{o}s-Renyi DAGs.

The reason for this is as follows. First note that for any trail of length at least $7$, there are at least $3$ intermediate nodes on the trail that are not colliders; this follows because assuming that all edges are unidirectional, colliders cannot be adjacent to each other and therefore if there are $k$ intermediate nodes, at least $\lfloor k/2\rfloor$ of them are non-colliders. 
Next consider a trail of length at least $7$ between a pair of nodes $a$ and $b$ and a conditioning set $S\in \{v_1,\ldots, v_n\}\setminus \{a,b\}$ of size $n-4$. Since $\{v_1,\ldots, v_n\}\setminus (S\cup \{a,b\})$ only has size $2$, it must be that $S$ contains at least one of the three or more non-colliders on the trail and hence blocks the trail.

\textbf{Statement 2 (Trails of length at most $6$).} In an Erd\H{o}s-Renyi DAG, the expected number of trails of length $\leq 6$ between a specific pair of nodes is
at most $p(1+(pn)+(pn)^2+\ldots (pn)^6)\leq 7n^6p^7$ where the last inequality assumes $p\geq 1/n$. If $p=\Theta(1)/n$, corresponding to graphs with constant expected degree, this expectation tends to 0 as $n\rightarrow \infty$ and hence, by the Markov bound, the probability there exists such a trail tends to $0$ as $n\rightarrow \infty$.

% $\frac{c}{n}+\frac{c^2}{n}+\frac{c^3}{n}+...+\frac{c^5}{n}$ where the probability of an edge existing is $\frac{c}{n}$.
% When $c=1$, this number is small.  

This statement can be proved as follows. There are ${n-2 \choose k}$ choices for the set of $k$ intermediate nodes on a trail of length $k+1$, there are $k!$ orderings of these nodes, and the probability there exists edges between each successive pair of nodes is $p^{k+1}$. Hence, the expected number of trails with $k$ intermediate nodes is $ k!{n-2 \choose k}p^{k+1}<n^k p^{k+1}$. Summing from $k=0$ to $6$ gives the claimed result.

% Simplify to get: $\leq n^k * p^{k+1}$. If $p=\frac{c}{n}$, then the expected number of trails with $k$ intermediate nodes is $\leq \frac{c^k}{n}$. Therefore, to compute the expected number of trails of length $2, 3, 4, 5, $ or $6$, add $\frac{c^k}{n}$ for $k=1, 2, 3, 4, 5$, from which we get: $\frac{c}{n}+\frac{c^2}{n}+\frac{c^3}{n}+...+\frac{c^5}{n}$. 

%Finally, we show that even when a  trail of length at most $6$ exists, the expected number of nodes that are colliders for some trail is very small. 

Finally we show that we expect most nodes to not be colliders for any trail. This implies that even when there are short trails, they will likely contain more non-colliders than colliders. And if the conditioning set includes at least one non-collider in a trail, then that trail is blocked. 

\textbf{Statement 3 (Expected number of colliders).} The expected number of nodes that are colliders for some trail is \[
% \sum_{i=1}^{n-2} (1-(1-p)^{n-i}-(n-i)*p*(1-p)^{n-i})=
n(1-p)^n-(1-p)^n+2((1-p)^n-1)/p+n-p^2+1 \ .\] 
% This is very small; 
For example, when $p={1}/{n}$, the expression corresponds to $0.104 n$. 

This is because the probability that $v_i$, the node with the $i$th largest label, is a collider is \begin{equation}
P(\textnormal{Bin}(i-1,p)\geq2)=1-(1-p)^{i-1}-(i-1)p(1-p)^{i-2}    \label{eq:binsum}
\end{equation} since the only way for a node $v_i$ to be a collider on any trail is if are edges between it and two nodes of more nodes amongst $\{v_1, \ldots v_{i-1}\}$.
% that have smaller topological labels than it. From the same reasoning, the probability that $v_{n-i}$ is a collider is $P(\textnormal{Bin}(n-i-1,p) \geq 2)$. 
Using linearity of expectation, the expected number of nodes that are colliders for some trail is 
% $E(\# \textnormal{ nodes that are colliders for some trail})=
$\sum_{i=3}^{n}
P(\textnormal{Bin}(i-1,p)\geq2)$
% (1-(1-p)^{n-i}-(n-i)*p*(1-p)^{n-i})$
and this simplifies to 
Eq.~\ref{eq:binsum}.
% =n*(1-p)^n-(1-p)^n+\frac{2*((1-p)^n-1)}{p}+n-p^2+1$.
% , which is very small.

\textbf{Takeaway.} For a single pair of variables, one unblocked trail will prevent the pre-processing algorithm from correctly determining that the pair could be made CI. Yet, we show that one unblocked trail of length $7$ or more is impossible, that the expected number of trails of length $2, 3, 4, 5,$ or $6$ is very small, and that the expected number of nodes that are colliders for a trail is very small. Since the expected number of short trails and of colliders is small, the large conditioning sets will very likely catch some pairs that would otherwise require many tests before the original PC algorithm would declare them to be conditionally independent.

\section{Experiments}

We performed experiments to estimate the raw number of CI tests performed by P3PC and by the original PC algorithm. Our goal was to evaluate the effects of pre-processing on the total number of tests run by the algorithms. Thus, we conducted experiments using known DAG structure and standard rules of d-separation to determine whether variables could be made conditionally independent by a given conditioning set, rather than running conditional independence tests on data generated by that DAG.

\begin{figure}[H]
\begin{center}
  	\includegraphics[scale=.85]{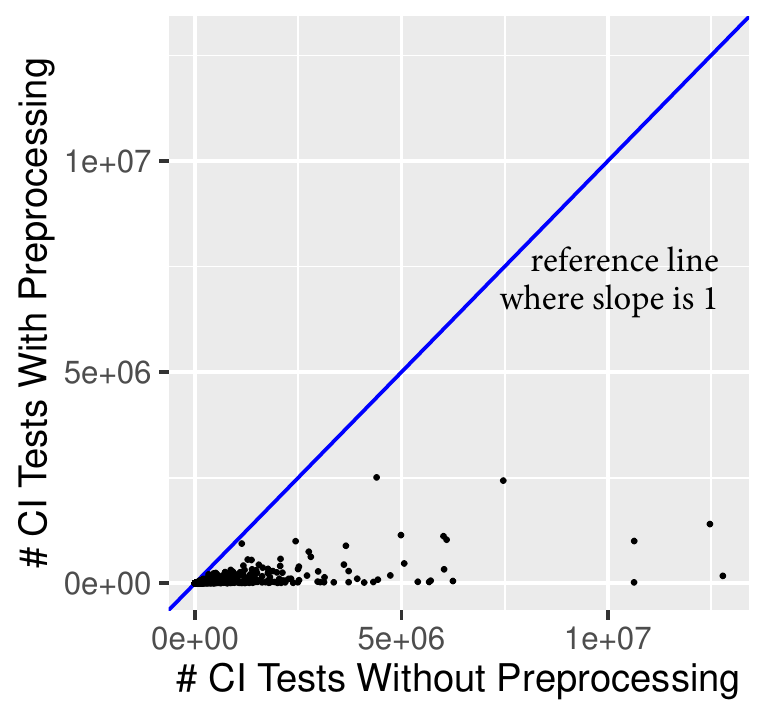}    
\end{center}
\caption{The number of tests performed by the PC algorithm with pre-processing (P3PC) including the number of tests performed in the pre-processing step on the y-axis and the number of tests performed by the PC algorithm without pre-processing on the x-axis. Each point in the plot corresponds to a randomly generated ER DAG and its location depends on the number of tests performed by both versions of the PC algorithm, with and without pre-processing. The blue line is for reference and has a slope of 1.}
\end{figure}

The first set of experiments compared the raw number of tests performed by the PC algorithm with and without pre-processing on over 1,000 Erd\H{o}s-Renyi DAGs randomly generated in the manner described in Section 4. The second set of experiments compared the same two algorithms on DAGs in the \texttt{bnlearn} repository, which are designed to correspond to real-world systems. For example, the \textit{Water} DAG is from a technical report on an expert system for controlling waste water treatment \cite{jensen1989}. The number of nodes in the Erd\H{o}s-Renyi DAGs is smaller than that of the realistic DAGs in the \texttt{bnlearn} repository because we aimed to perform experiments on a large number of ER DAGs and inferring d-separation is time-consuming on larger, dense DAGs.

In both the DAGs that correspond to real-world systems and the Erd\H{o}s-Renyi DAGs, the raw number of tests performed by P3PC is much smaller than those performed by the PC algorithm alone as shown in Figure 1. The efficiency gain is greater as the number of edges increases to be twice the amount of the number of nodes. The difference in performance of P3PC and the PC algorithm alone is more noticeable for the DAGs that correspond to real-world systems, where the number of tests performed by P3PC is so much smaller than that performed by the PC algorithm alone that the bar representing that number often is barely visible in Figure 2. Specifically, for the DAGs that correspond to real-world systems, the number of tests performed by P3PC is between 0.5\% to 36\%, and often less than 10\%, of the number of tests performed by the PC algorithm alone as displayed in Table 1. %Further, the difference is starker when the number of CI tests that the PC algorithm without pre-processing performs is much higher. 

{
\footnotesize
\begin{table}[H]
\small
\begin{tabular}{ c|c c c c c c c c c c } 
 &  &  &  &  &  &  &  & Magic &  & Magic \\
 & Child & Alarm & Mildew & Ecoli & Insurance & Water & Barley & Niab & Mehra & Irri \\ 
\hline
 \# Nodes & 20 & 37 & 35 & 46&27 & 32& 48& 44& 24& 64 \\ 
 \# Edges & 25 & 46 & 46 & 70&52 & 66& 84& 66& 71& 102\\ 
 Prop. Tests & 0.361 & 0.143 & 0.055 & 0.285&0.073 & 0.021& 0.006& 0.008& 0.137& 0.005\\ 
 \hline
\end{tabular}~\\~\\
\caption{\hspace{0.1in} The number of nodes and edges for each DAG from the \texttt{bnlearn} repository that corresponds to real-world systems, and the proportion of tests performed by P3PC to tests performed by the PC algorithm alone for each.}
\end{table}}

\begin{figure}[H]
	%\begin{subfigure}{0.31\linewidth}
  %		\includegraphics[scale=.24]{child.JPG}
%	\end{subfigure}
%	\begin{subfigure}{0.2\linewidth}
%		\includegraphics[scale=.16]{insurance.JPG}
%	\end{subfigure}
%	\begin{subfigure}{0.22\linewidth}
%		\includegraphics[scale=.17]{barley.JPG}
%	\end{subfigure}
%	\begin{subfigure}{0.2\linewidth}
%		\includegraphics[scale=.16]{mehra.JPG}
%	\end{subfigure}
\begin{center}
  	\includegraphics[scale=.35]{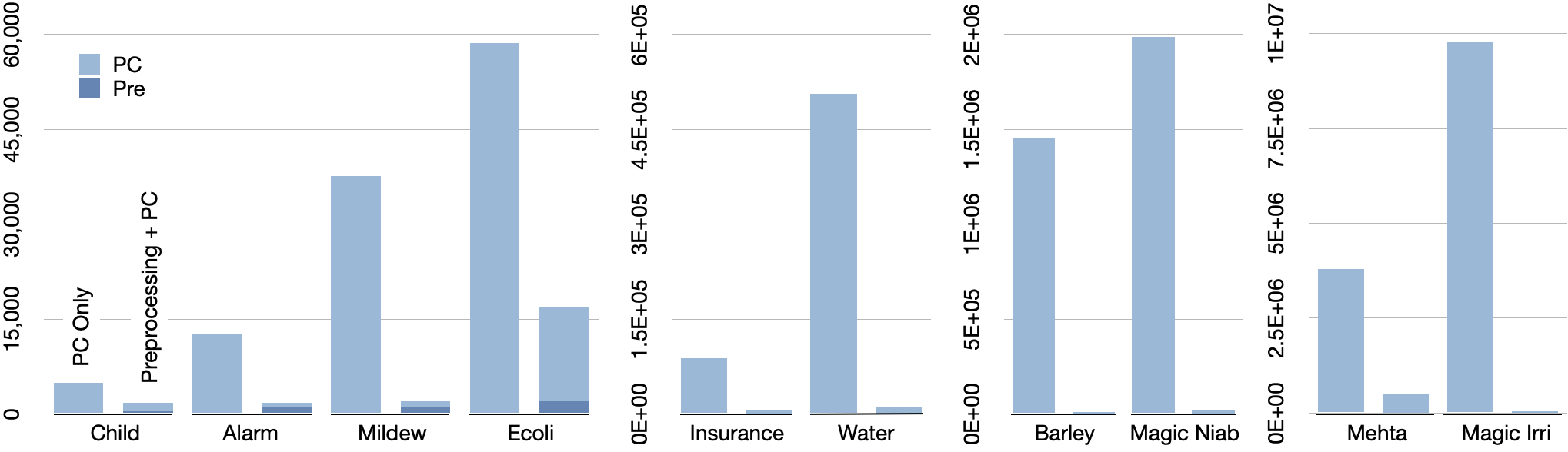}
\end{center}
\caption{For DAGs that correspond to real-world systems in the \texttt{bnlearn} repository, the number of tests that P3PC performs to learn a structure and that the PC algorithm alone performs to learn a structure. The x-axis of the plots are the names of realistic DAGs, while the y-axis of the plots are the number of tests performed. Note that the y-axis varies for different DAGs. For all realistic DAGs, the raw number of tests performed by P3PC is much smaller than that performed by the PC algorithm alone.}

\end{figure}

\section{Conclusion}

We proposed a pre-processing algorithm for the PC algorithm which can dramatically reduce the total number of CI tests that the PC algorithm performs. This implies a new reason to prioritize research on accurate model-based tests of conditional independence, particularly those that perform well with large conditioning sets. Furthermore, we show theoretically why the pre-processing algorithm reduces this number in expectation for Erd\H{o}s-Renyi graphs. Finally, we perform experiments on DAGs that correspond to real-world systems, finding that the number of tests performed by P3PC is between 0.5\% to 36\%, and often less than 10\%, of the number of tests performed by the PC algorithm alone. From the experiments on Erd\H{o}s-Renyi graphs, we observe that the efficiency gain is also large and increases as the number of edges increases to be twice the number of nodes.

This work addresses one of the primary challenges for routine use of the PC algorithm by greatly reducing runtime. This makes possible improved algorithms such as PC-stable, which repeats the steps in the PC algorithm many times, more tractable \cite{colombo2014order}. To further increase the efficiency of the pre-processing step, a system could parallelize the tests within the pre-processing algorithm for each pair since the tests for different pairs do not depend on each other. \raggedbottom

\section*{Acknowledgments}

Thanks to Kaleigh Clary and Purva Pruthi for their helpful feedback and suggestions. This research was sponsored by the National Science Foundation, the Defense Advanced Research Projects Agency (DARPA), the United States Air Force, and the Army Research Office (ARO) under grants NSF CCF-1934846 and NSF CCF-1908849, Contract Number FA8750-17-C-0120, and Cooperative Agreements Number W911NF-20-2-0005 and W911NF-20-2-0005. The views and conclusions contained in this document are those of the authors and should not be interpreted as representing the official policies, either expressed or implied, of NSF, DARPA, the USAF, ARO, or the U.S. Government. The U.S. Government is authorized to reproduce and distribute reprints for Government purposes not withstanding any copyright notation herein.

%\pagebreak
%\bibliography{ref}

\end{document}